\documentclass[runningheads,a4paper]{llncs}
\pdfoutput=1 

\usepackage[utf8]{inputenc}
\usepackage{amssymb}
\usepackage{graphicx}
\usepackage[ruled,vlined]{algorithm2e}
\usepackage{amsmath}
\usepackage{times}
\usepackage{array}
\usepackage{graphicx}
\usepackage{subfigure} 
\usepackage{natbib}
\usepackage{url}

\newcommand{\keywords}[1]{\par\addvspace\baselineskip
  \noindent\keywordname\enspace\ignorespaces#1}

\begin{document}

\mainmatter

\title{Sampled Weighted Min-Hashing \\
  for Large-Scale Topic Mining}

\titlerunning{Sampled Weighted Min-Hashing for Large-Scale Topic Mining}

\author{Gibran Fuentes-Pineda
  \and Ivan Vladimir Meza-Ruiz}
\authorrunning{Fuentes-Pineda and Meza-Ruiz}

\institute{Instituto de Investigaciones en Matemáticas y en Sistemas\\ Universidad Nacional Autónoma de México}

\maketitle

\begin{abstract}
  We present Sampled Weighted Min-Hashing (SWMH), a randomized approach to 
  automatically mine topics from large-scale corpora. SWMH generates multiple 
  random partitions of the corpus vocabulary based on term co-occurrence and 
  agglomerates highly overlapping inter-partition cells to produce the mined 
  topics. While other approaches define a topic as a probabilistic distribution 
  over a vocabulary, SWMH topics are ordered subsets of such 
  vocabulary. Interestingly, the topics mined by SWMH underlie themes from the 
  corpus at different levels of granularity. We extensively evaluate the meaningfulness of the mined topics
  both qualitatively and quantitatively on the NIPS (1.7K documents), 20 
  Newsgroups (20K), Reuters (800K) and Wikipedia (4M) corpora.  Additionally, we 
  compare the quality of SWMH with Online LDA topics for document representation in 
  classification.
\end{abstract}

\section{Introduction}
The automatic extraction of topics has become very important in recent years 
since they provide a meaningful way to organize, browse and represent large-scale collections of documents. 
Among the most successful approaches to topic discovery are directed topic models such as Latent Dirichlet Allocation (LDA)~\cite{lda} and 
Hierarchical Dirichlet Processes (HDP)~\cite{hdp} which are Directed Graphical Models with latent topic 
variables. More recently, 
undirected graphical models have been also applied to topic modeling, (e.g., Boltzmann Machines~\cite{Ruslan_undirected_nips2009,Srivastava_uai2013} and Neural Autoregressive Distribution 
Estimators~\cite{larochelle_nips2012}). The topics generated by both directed 
and undirected models have been shown to underlie the thematic structure of a text 
corpus. These topics are defined as distributions over terms of a 
vocabulary and documents in turn as distributions over topics. 
Traditionally, inference in topic models has not scale well to large corpora, however, more efficient strategies 
have been proposed to overcome this problem (e.g., Online LDA~\cite{onlinelda} and stochastic 
variational inference~\cite{Mimno_icml2012}). Undirected Topic Models can be 
also trained efficiently using approximate strategies such as Contrastive 
Divergence~\cite{condiver}. 

In this work, we explore the mining of topics based on term co-occurrence.
The underlying intuition is that terms consistently co-occurring in the same documents are likely to belong to the same topic.
The resulting topics correspond to ordered subsets of 
the vocabulary rather than distributions over such a vocabulary. 
Since finding co-occurring terms is a combinatorial problem that lies in a 
large search space, we propose Sampled Weighted Min-Hashing (SWMH), an extended version 
of Sampled Min-Hashing (SMH)~\cite{fuentes}.
SMH partitions the vocabulary into sets of highly co-occurring terms by applying Min-Hashing~\cite{minhash} to the inverted file entries of the corpus.
The basic idea of Min-Hashing is to generate random partitions of the space so that sets with high Jaccard similarity are more likely to lie in the same partition cell.

One limitation of SMH is that the generated random partitions are drawn from 
uniform distributions. This setting is not ideal for information 
retrieval applications where weighting have a positive impact on the quality of the 
retrieved documents~\cite{salton,buckley}. For this reason, we extend SMH by allowing
weights in the mining process which effectively extends the uniform 
distribution to a distribution based on weights. 
We demonstrate the validity and scalability of the proposed approach by mining topics in the NIPS, 
20 Newsgroups, Reuters and Wikipedia corpora which range from small (a thousand of documents) to large scale (millions of documents). 
Table~\ref{tbl:examples} presents some examples of mined topics and their sizes. 
Interestingly, SWMH can mine meaningful topics of different levels of granularity. 

\begin{table*}[t]
  \caption{SWMH topic examples.}
  \centering
  \begin{tabular}{|l|l|}\hline
    NIPS
    & 
    introduction,references,shown,figure,abstract,shows,back,left,process,\ldots 
    (51)\\
    & chip,fabricated,cmos,vlsi,chips,voltage,capacitor,digital,inherent,\ldots 
    (42)\\
    & spiking, spikes, spike,firing, cell, neuron, 
    reproduces,episodes,cellular,
    \ldots (17)\\
    \hline
    20 Newsgroups & algorithm communications clipper encryption chip key \\
    & lakers, athletics, alphabetical, pdp, rams, pct,  mariners, clippers, 
    \ldots (37)\\
    & embryo, embryos, infertility, ivfet, safetybelt, gonorrhea, dhhs, \ldots (37)\\\hline
    Reuters & prior, quarterly, record, pay, amount, latest, oct \\
    & precious, platinum, ounce, silver, metals, gold \\
    & udinese, reggiana, piacenza, verona, cagliari, atalanta, perugia, \ldots (64)\\
    \hline
    Wikipedia & median, householder, capita, couples, racial, makeup, residing,
    \ldots (54)  \\
    & decepticons', galvatron's, autobots', botcon, starscream's, rodimus, galvatron\\
    & avg, strikeouts, pitchers, rbi, batters, pos, starters, pitched, hr, batting, \ldots (21)\\
    \hline
  \end{tabular}
  \label{tbl:examples}
\end{table*}

The remainder of the paper is organized as follows. Section~\ref{sec:minhash} reviews the Min-Hashing scheme for pairwise set similarity search. The proposed approach for topic mining by SWMH is described in Sect.~\ref{sec:smh}. Section~\ref{sec:exp} reports the experimental evaluation of SWMH as well as a comparison against Online LDA. Finally, Sect.~\ref{sec:concl} concludes the paper with some discussion and future work.

\section{Min-Hashing for Pairwise Similarity Search}
\label{sec:minhash}
Min-Hashing is a randomized algorithm for efficient pairwise set similarity search (see Algorithm~\ref{algo:sim}). The basic idea is to define MinHash functions $h$ with the property that the probability of any two sets $A_1, A_2$ having the same MinHash value is equal to their Jaccard Similarity, i.e.,

\begin{equation}
  \label{eq:mh_prob}
  P[h(A_1) = h(A_2)] = \frac{\mid A_1 \cap A_2 \mid}{\mid A_1 \cup A_2 \mid}
  \in [0,1].
\end{equation}

Each MinHash function $h$ is realized by generating a random permutation $\pi$ of all the elements 
and assigning the first element of a set on the permutation as its MinHash value. 
The rationale behind Min-Hashing is that similar sets will have a high probability of taking the same MinHash value whereas dissimilar sets will have a low probability.
To cope with random fluctuations, multiple MinHash values are computed for each set from independent random permutations. 
Remarkably, it has been shown that the portion of identical MinHash values between two sets is an unbiased estimator of their Jaccard similarity~\cite{minhash}.

Taking into account the above properties, in Min-Hashing similar sets are retrieved by grouping $l$ tuples $g_1, \ldots, g_l$ of $r$ different MinHash values as follows

\begin{equation*}
  \begin{tabular}{l}
    $g_1(A_1) = (h_1(A_1), h_2(A_1), \ldots , h_r(A_1))$\\
    $g_2(A_1) = (h_{r+1}(A_1), h_{r+2}(A_1), \ldots , h_{2\cdot r}(A_1))$\\
    $\cdots$ \\
    $g_l(A_1) = (h_{(l-1)\cdot r+1}(A_1), h_{(l-1)\cdot r+2}(A_1), \ldots , h_{l\cdot r}(A_1))$
  \end{tabular},
\end{equation*}

\noindent where $h_j(A_1)$ is the $j$-th MinHash value. 
Thus, $l$ different hash tables are constructed and two sets 
$A_1, A_2$ are stored in the same hash bucket on the $k$-th hash table if 
$g_k(A_1) = g_k(A_2), k = 1, \ldots , l$. Because similar sets are expected to agree in several MinHash values, they will be stored in the same hash bucket with high probability. 
In contrast, dissimilar sets will seldom have the same MinHash value and therefore the 
probability that they have an identical tuple will be low. More precisely, the probability 
that two sets $A_1,A_2$ agree in the $r$ MinHash values of a given tuple $g_k$ is $P[g_k(A_1) = g_k(A_2)] = sim(A_1,A_2)^r$. Therefore, the probability that two sets $A_1,A_2$ have at least one identical tuple is  $P_{collision}[A_1,A_2] = 1-(1-sim(A_1,A_2)^r)^l$.

The original Min-Hashing scheme was extended by Chum et al.~\cite{chum_wminhash} to weighted set similarity, defined as 

\begin{equation}
  sim_{hist}(H_1, H_2) = \frac{\sum_i w_i \min (H_1^i, H_2^i)}{\sum_i w_i \max(H_1^i, H_2^i)} \in [0,1],
\end{equation}

\noindent where $H_1^i, H_2^i$ are the frecuencies of the $i$-th element in the histograms $H_1$ and $H_2$ respectively and $w_i$ is the weight of the element. In this scheme, instead of generating random permutations drawn from a uniform distribution, the permutations are drawn from a distribution based on element weights. This extension allows the use of popular document representations based on weighting schemes such as \emph{tf-idf} and has been applied to image retrieval~\cite{chum_wminhash} and clustering~\cite{chum2010}.

\begin{algorithm}[tb]
  \SetAlgoLined
  \KwData{Database of sets $A = A_1, \ldots , A_N$ and query set $q$}
  \KwResult{Similar sets to $q$ in $A$}
           {\bf Indexing}
           \begin{enumerate}
           \item Compute $l$ MinHash tuples $g_i(A_j), i = 1, \ldots , l$ for each set $A_j, j = 1, \ldots, N$ in $A$.
           \item Construct $l$ hash tables and store each set $A_j, j = 1, \ldots, N$ in the buckets corresponding to $g_i(A_j), i = 1, \ldots , l$.
           \end{enumerate}
           
           {\bf Querying}
           \begin{enumerate}
           \item Compute the $l$ MinHash tuples $g_i(q), i = 1, \ldots , l$ for the query set $q$.
           \item Retrieve the sets stored in the buckets corresponding to $g_i(q), i = 1, \ldots , l$.
           \item Compute the similarity between each retrieved set and $q$ and return those with similarity greater than a given threshold $\epsilon$.
           \end{enumerate}
           \caption{Pairwise Similarity Search by Min-Hashing}
           \label{algo:sim}
\end{algorithm}

\section{Sampled Min-Hashing for Topic Mining}
\label{sec:smh}
Min-Hashing has been used in document and image retrieval and classification, where documents and images are represented as bags of words. Recently, it was also successfully applied to retrieving co-occurring terms by hashing the inverted file lists instead of the documents~\cite{chum_wminhash,fuentes}.
In particular, Fuentes-Pineda et al.~\cite{fuentes} proposed Sampled Min-Hashing (SMH), a simple strategy based on Min-Hashing to discover objects from large-scale image collections. In the following, we briefly describe SMH using the notation of terms, topics and documents, although it can be generalized to any type of dyadic data. The underlying idea of SMH is to mine groups of terms with high \emph{Jaccard Co-occurrence Coefficient (JCC)}, i.e.,  

\begin{equation}
  \label{eq:jcc}
  JCC(T_1, \ldots , T_k) = \frac{\vert T_1 \cap T_2 \cap \cdots \cap T_k \vert }{\vert T_1 \cup T_2 \cup \cdots \cup T_k \vert},
\end{equation}

\noindent where the numerator correspond to the number of documents in which terms $T_1, \ldots ,T_k$ co-occur
and the denominator is the number of documents with at least one of the $k$ terms. Thus, Eq.~\ref{eq:mh_prob} can be extended to multiple co-occurring terms as

\begin{equation}
  \label{eq:jcc_prob}
  P[h(T_1) = h(T_2) \ldots = h(T_k)] = JCC(T_1, \ldots , T_k).
\end{equation}

From Eqs.~\ref{eq:jcc} and~\ref{eq:jcc_prob}, it is clear that the probability that all terms $T_1, \ldots , T_k$ have the same MinHash value 
depends on how correlated their occurrences are: the more correlated the higher is the probability of taking the same MinHash value.  
This implies that terms consistently co-occurring in many documents will have a high probability of taking the same MinHash value.

In the same way as pairwise Min-Hashing, $l$ tuples of $r$ MinHash values are computed to find groups of terms with identical tuple, which become a co-occurring term set. 
By choosing $r$ and $l$ properly, the probability that a group of $k$ terms has an identical tuple approximates a unit step function such that

\begin{equation*}
  P_{collision}[T_1, \ldots , T_k] \approx
  \begin{cases}
    1 & \mbox{if } JCC(T_1, \ldots , T_k) \geq s* \\
    0 & \mbox{if } JCC(T_1, \ldots , T_k) < s*
  \end{cases},
\end{equation*}

Here, the selection of $r$ and $l$ is a trade-off between precision and recall. 
Given $s*$ and $r$, we can determine $l$ by setting $P_{collision}[T_1, \ldots ,T_k]$ to $0.5$, which gives 

\begin{equation*}
  l = \frac{\log(0.5)}{\log(1 - s*^r)}
\end{equation*}.

\begin{figure}[t]
  \centering
  \includegraphics[scale=0.24]{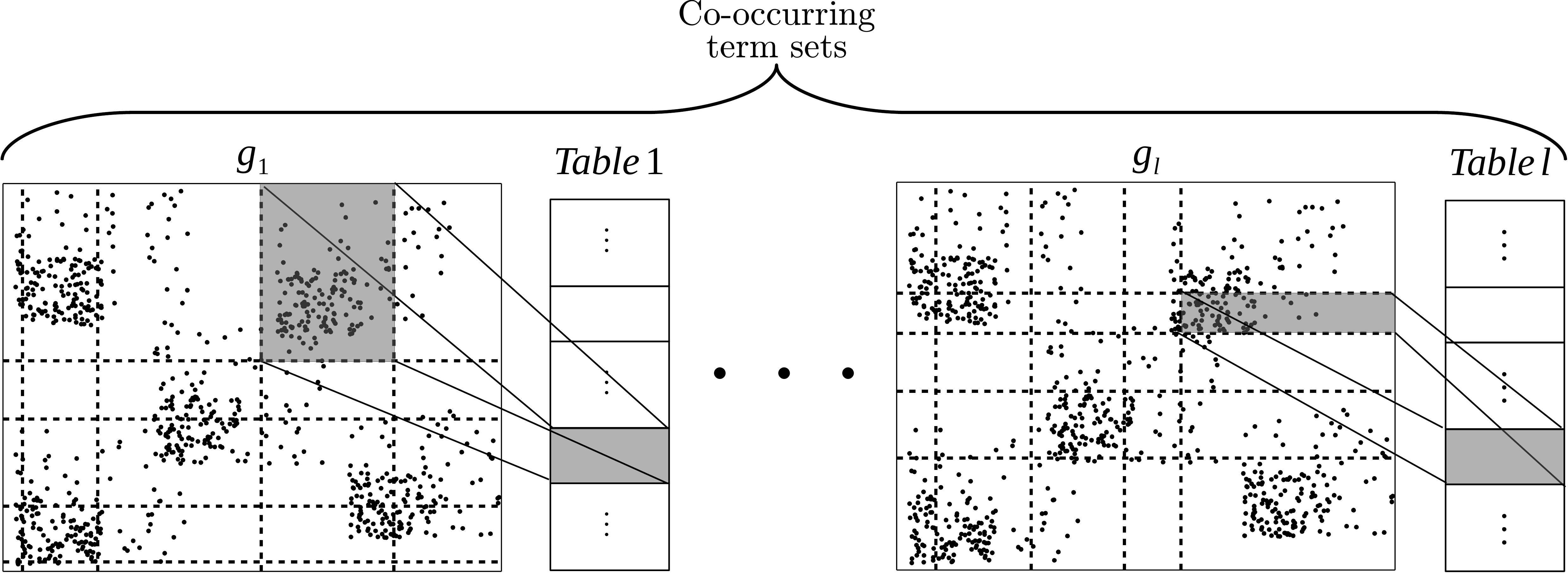}
  \caption[]{Partitioning of the vocabulary by Min-Hashing.}
  \label{fig:rand_part}
\end{figure}

In SMH, each hash table can be seen as a random partitioning of the vocabulary into disjoint groups of highly co-occurring terms, as illustrated in Fig.~\ref{fig:rand_part}. Different partitions are generated and groups of discriminative and stable terms belonging to the same topic are expected to lie on overlapping inter-partition cells.
Therefore, we cluster co-occurring term sets that share many terms in an agglomerative manner.
We measure the proportion of terms shared between two co-occurring term sets $C_1$ and $C_2$ by their overlap coefficient, namely

\begin{equation*}
  ovr(C_1,C_2) = \frac{\mid C_1 \cap C_2 \mid}{\min(\mid C_1 \mid,\mid C_2\mid )} \in [0,1].
\end{equation*}

\noindent Since a pair of co-occurring term sets with high Jaccard similarity will 
also have a large overlap coefficient, finding pairs of co-occurring term sets can 
be speeded up by using Min-Hashing, thus avoiding the overhead of computing the overlap coefficient between all the pairs of co-occurring term sets.

The clustering stage merges chains of co-occurring term sets with high overlap 
coefficient into the same topic.
As a result, co-occurring term sets associated with the same topic can belong 
to the same cluster even if they do not share terms with one another, as 
long as they are members of the same chain.
In general, the generated clusters have the property that for any co-occurring 
term set, there exists at least one co-occurring term set in the 
same cluster with which it has an overlap coefficient greater than a given threshold $\epsilon$. 

We explore the use of SMH to mine topics from documents but we judge term co-occurrence by the Weighted Co-occurrence Coefficient (WCC), defined as

\begin{equation}
  \mathit{WCC}(T_1, \ldots , T_k) = \frac{\sum_i w_i \min{(T_1^i, \cdots , T_k^i)}}{\sum_i w_i \max{(T_1^i, \cdots , T_k^i )}} \in [0,1], 
\end{equation}

\noindent where $T_1^i, \cdots , T_k^i$ are the frecuencies in which terms $T_1, \ldots , T_k$ occur in the $i$-th document and the weight $w_i$ is given by the inverse of the size of the $i$-th document. We exploit the extended Min-Hashing scheme by Chum et al.~\cite{chum_wminhash} to efficiently find such co-occurring terms. We call this topic mining strategy Sampled Weighted Min-Hashing (SWMH) and summarize it in Algorithm~\ref{algo:swmh}. 

\begin{algorithm}[tb]
  \SetAlgoLined
  \KwData{Inverted File Lists $T = T_1, \ldots , T_N$}
  \KwResult{Mined Topics $O = O_1, \ldots , O_M$}
           {\bf Partitioning}
           \begin{enumerate}
           \item Compute $l$ MinHash tuples $g_i(T_j), i = 1, \ldots , l$ for each list $T_j, j = 1,\ldots, N$ in $T$.
           \item Construct $l$ hash tables and store each list $T_j, j = 1,\ldots, N$ in the bucket corresponding to $g_i(T_j), i = 1, \ldots , l$.
           \item Mark each group of lists stored in the same bucket as a co-occurring term set.
           \end{enumerate}
           
           {\bf Clustering}
           \begin{enumerate}
           \item Find pairs of co-occurring term sets with overlap coefficient greater than a given threshold $\epsilon$.
           \item Form a graph $G$ with co-occurring term sets as vertices and edges defined between pairs with overlap coefficient greater than $\epsilon$.
           \item Mark each connected component of $G$ as a topic.
           \end{enumerate}
           
           \caption{Topic mining by SWMH}
           \label{algo:swmh}
\end{algorithm}

\section{Experimental Results}
\label{sec:exp}
In this section, we evaluate different aspects of the mined topics. First, we 
present a comparison between the topics mined by SWMH and SMH. Second, we 
evaluate the scalability of the proposed approach. Third, we use the 
mined topics to perform document classification. Finally, we compare SWMH 
topics with Online LDA topics. 

The corpora used in our experiments were: NIPS, 20 Newsgroups, Reuters and 
Wikipedia\footnote{Wikipedia dump from 2013--09--04.}. NIPS is a small 
collection of articles ($3,649$ documents), 20 Newsgroups is a larger 
collection of mail newsgroups ($34,891$ documents), Reuters is a medium size 
collection of news ($137,589$ documents) and Wikipedia is a large-scale 
collection of encyclopedia articles ($1,265,756$ documents) \footnote{All 
  corpora were preprocessed to cut off terms that appeared less than 6 times in 
  the whole corpus.}. 

All the experiments presented in this work were performed on an Intel(R) Xeon(R) 2.66GHz workstation with 8GB of memory and with 8 processors. 
However, we would like to point out that the current version of the code is not parallelized, so we did not take advantage of the multiple
processors.

\subsection{Comparison between SMH and SWMH}
\begin{figure}[t]
  \centering
  \includegraphics[scale=0.336]{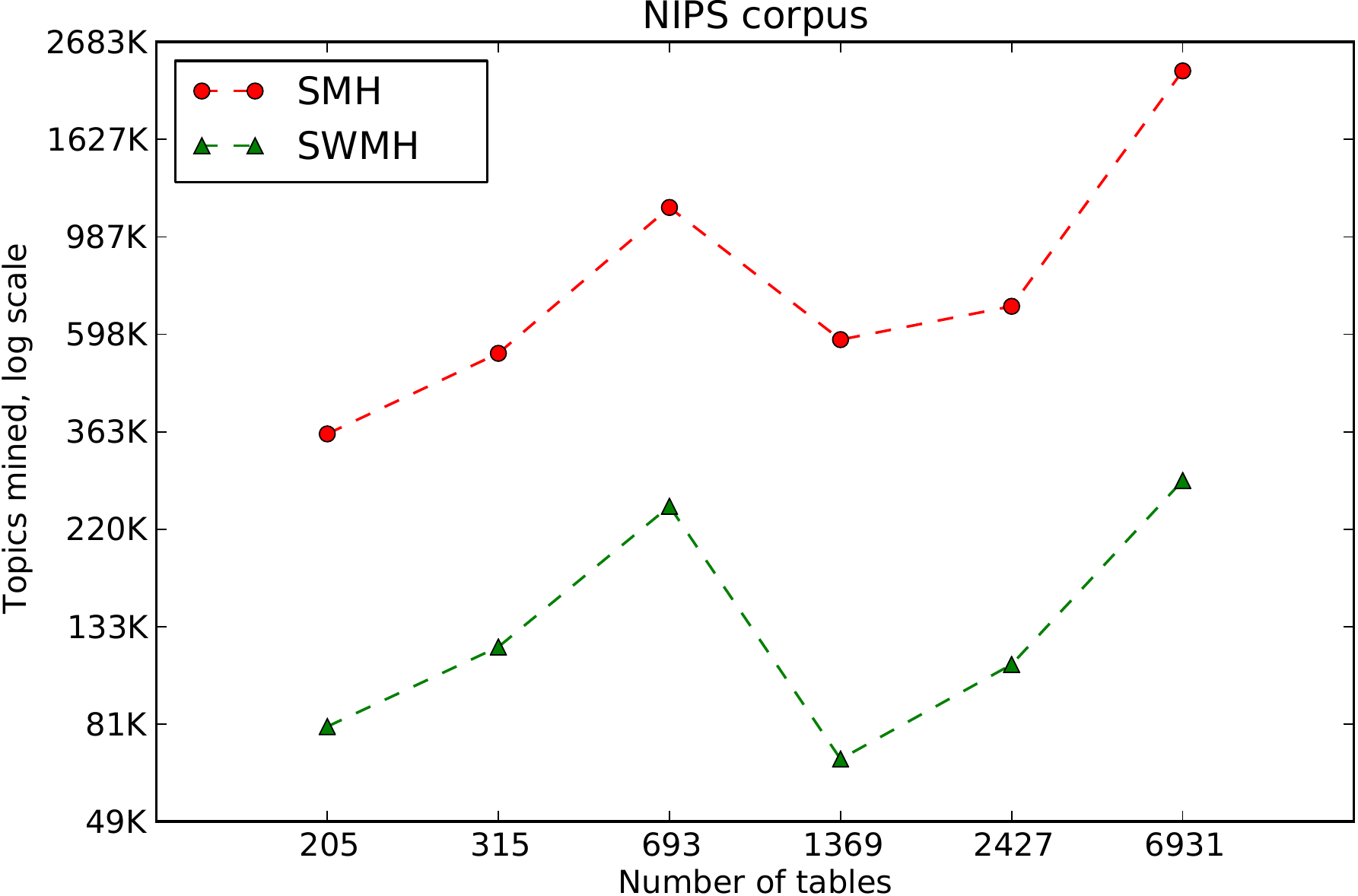}
  \includegraphics[scale=0.336]{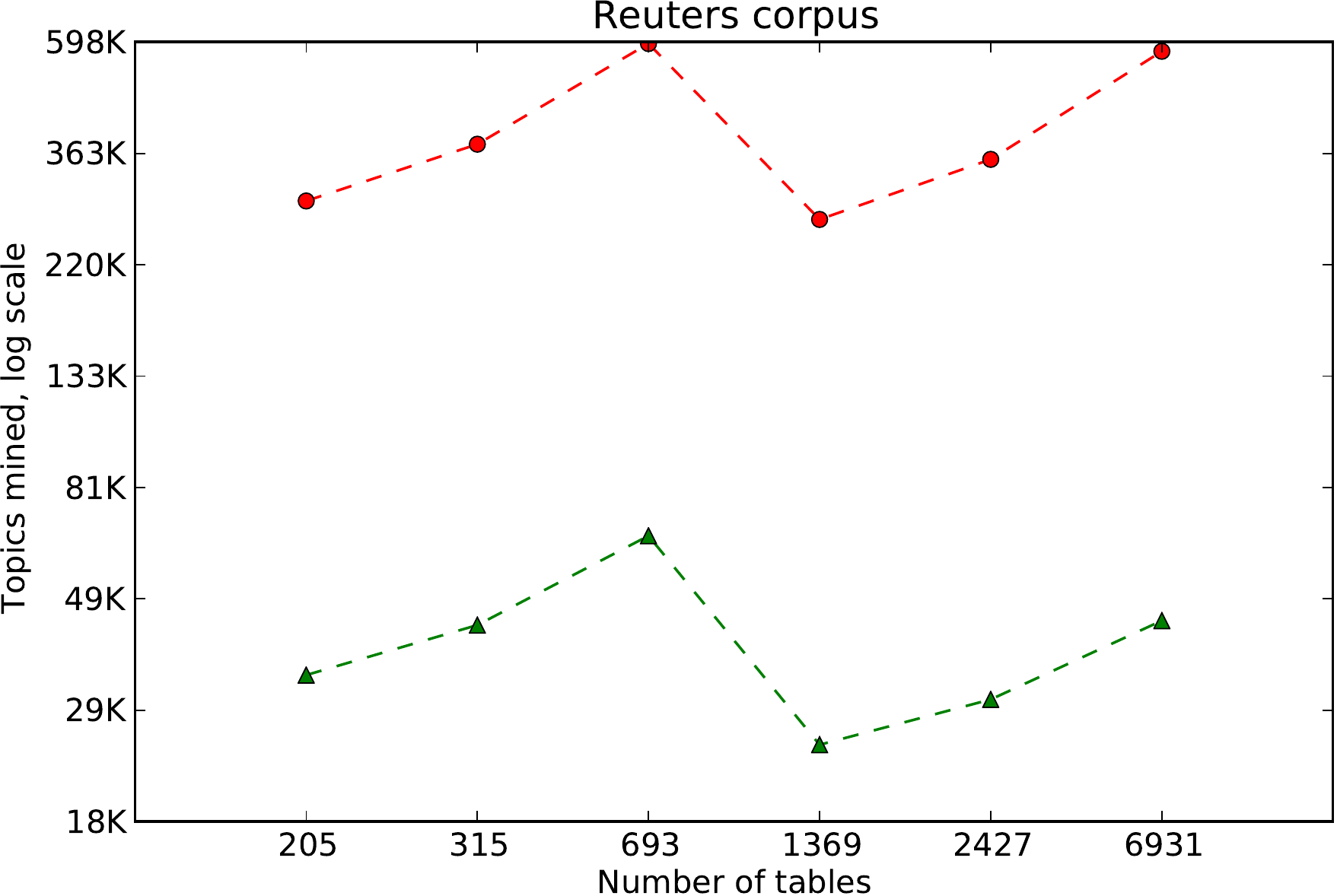}
  \caption[]{Amount of mined topics for SMH and SWMH in the (a) NIPS and (b) Reuters corpora.}
  \label{fig:sizes}
\end{figure}

For these experiments, we used the NIPS and 
Reuters corpora and different values of the parameters $s*$ and $r$, which define the number of MinHash tables.
We set the parameters of 
similarity ($s*$) to $0.15$, $0.13$ and $0.10$ and the tuple size ($r$) to $3$ 
and $4$. These parameters rendered the following table sizes: $205$, $315$, 
$693$, $1369$, $2427$, $6931$. Figure \ref{fig:sizes} shows the effect of 
weighting on the amount of mined topics.  First, notice the breaking point on 
both figures when passing from $1369$ to $2427$ tables.  This effect 
corresponds to resetting the $s*$ to $.10$ when changing $r$ from $3$ to $4$.  
Lower values in $s*$ are more strict and therefore less topics are mined.  
Figure \ref{fig:sizes} also shows that the amount of mined topics is 
significantly reduced by SWMH, since the colliding terms not 
only need to appear on similar documents but now with similar proportions. The 
effect of using SWMH is also noticeable in the number of terms that compose a 
topic. The maximum reduction reached in NIPS was $73\%$ while in Reuters was 
$45\%$. 

\subsection{Scalability evaluation}
\label{sec:lse}
\begin{figure}[t]
  \centering
  \includegraphics[scale=0.25]{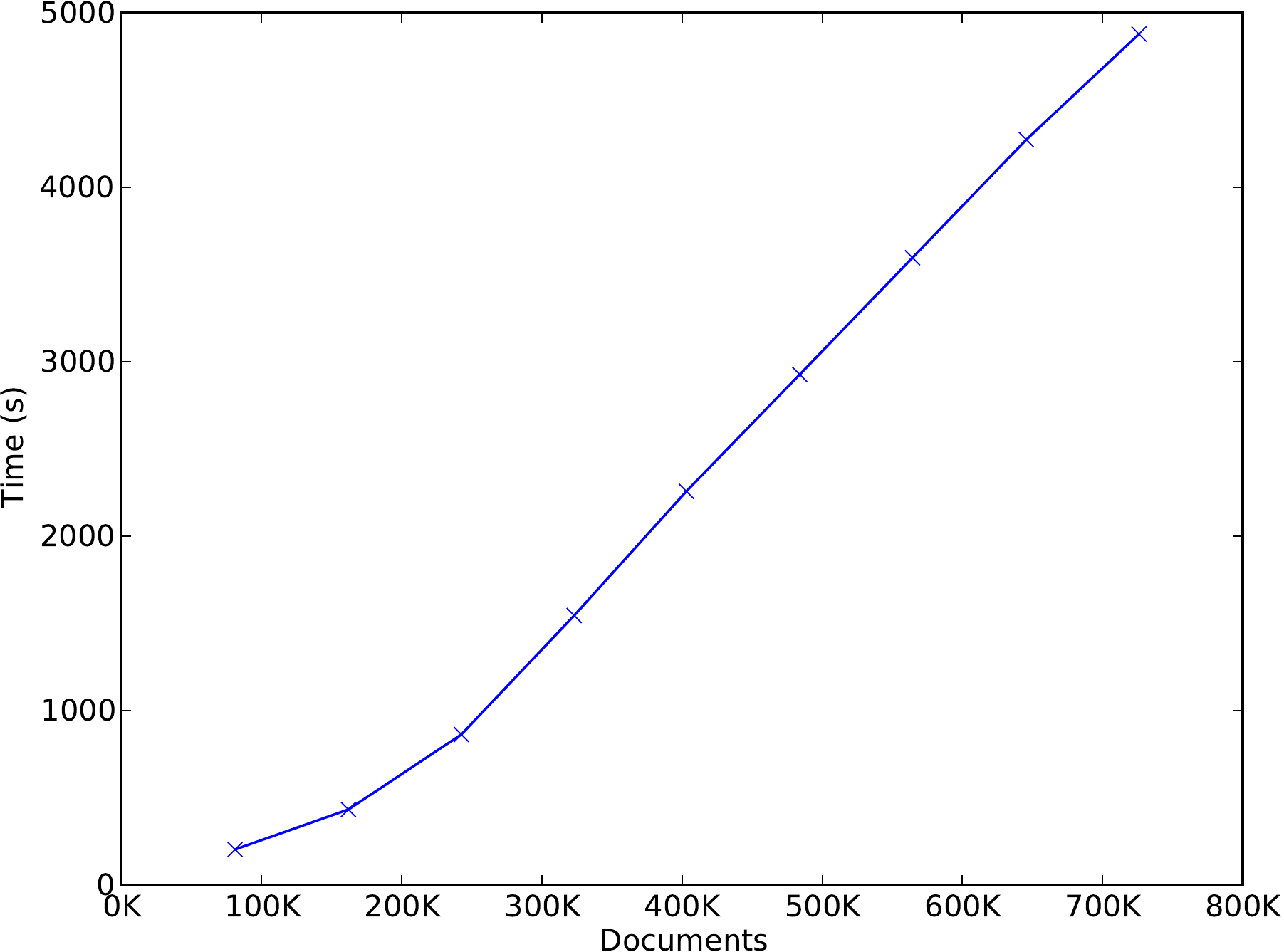}\hspace{1cm}
  \includegraphics[scale=0.25]{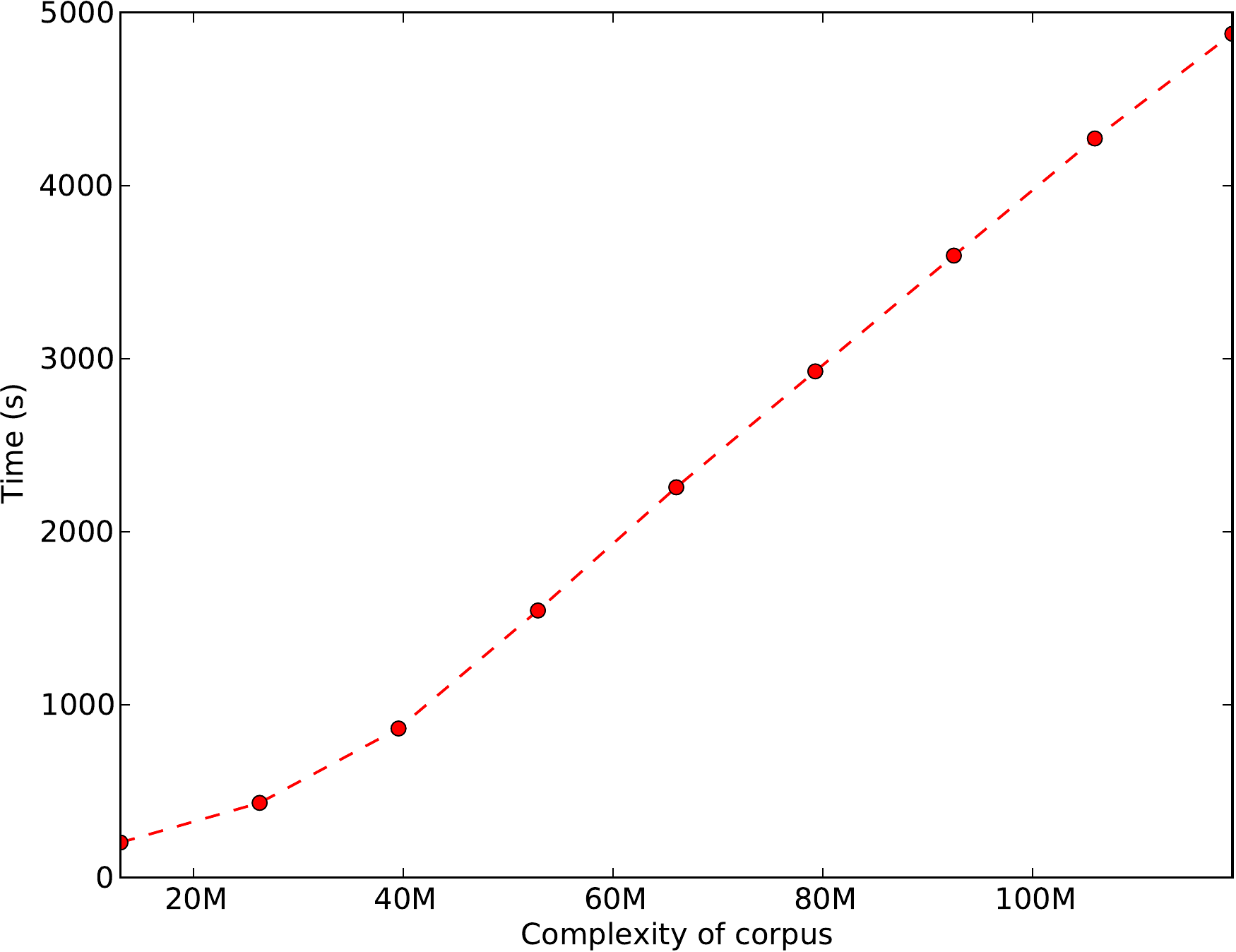}
  \caption[]{Time scalability for the Reuters corpus.}
  \label{fig:scalability}
\end{figure}

To test the scalability of SWMH, we measured the time and memory required to mine 
topics in the Reuters corpus while increasing the number of documents to be analyzed. 
In particular, we perform 10 experiments with SWMH, each increasing the number of documents by 10\% \footnote{The parameters were fixed to $s*=0.1$,$r=3$, 
  and overlap threshold of $0.7$.}. Figure~\ref{fig:scalability} illustrates the time taken to mine topics as we increase the number of documents and as we increase an index of complexity given by a combination of the size of the vocabulary and the average number of times a term appears in a document. As can be noticed, in both cases the time grows almost 
linearly and is in the thousand of seconds.

The mining times for the corpora were: NIPS, $43s$; 20 Newsgroups, $70s$; Reuters, $4,446s$ and Wikipedia, $45,834s$. These times contrast with the required time by Online LDA to model 100 topics \footnote{\url{https://github.com/qpleple/online-lda-vb} was adapted 
  to use our file formats.}: NIPS, $60s$; 20 Newsgroups, $154s$ and  Reuters, $25,997$. Additionally, we set Online LDA to model $400$ topics with the Reuters corpus and took $3$ days. Memory figures follow a similar behavior to the time figures. Maximum memory: NIPS, $141MB$; 20 Newsgroups, $164MB$; Reuters, $530MB$ and Wikipedia, $1,500MB$.

\subsection{Document classification}
In this evaluation we used the mined topics to create a document representation 
based on the similarity between topics and documents. This 
representation was used to train an SVM classifier with the class of the 
document. In particular, we focused on the 20 Newsgroups corpus for this 
experiment. We used the typical setting of this corpus for document 
classification ($60\%$ training, $40\%$ testing). Table \ref{tbl:docclass} 
shows the performance for different variants of topics mined by SWMH and Online LDA topics. 
The results illustrate that the number of topics is relevant for the 
task: Online LDA with $400$ topics is better than $100$ topics. A similar 
behavior can be noticed for SWMH, however, the parameter $r$ has an 
effect on the content of the topics and therefore on the performance.

\begin{table}[t]
  \centering
  \caption{Document classification for 20 Newsgroups corpus.} 
  \begin{tabular}{|l|c|c|c|}\hline
    Model	   & Topics  & Accuracy      & Avg. score  \\\hline
    $205$      & $3394$ &  $59.9$       & $60.6$   \\
    $319$      & $4427$ &  $61.2$       & $64.3$   \\
    $693$      & $6090$ &  $68.9$       & $70.7$   \\
    $1693$     & $2868$ &  $53.1$       & $55.8$   \\
    $2427$     & $3687$ &  $56.2$       & $60.0$   \\
    $6963$     & $5510$ &  $64.1$       & $66.4$   \\\hline
    Online LDA & $100$ &  $59.2$       & $60.0$   \\
    Online LDA & $400$ &  $65.4$       & $65.9$   \\\hline
  \end{tabular}
  \label{tbl:docclass}
\end{table}

\subsection{Comparison between mined and modeled topics}

\begin{figure}[t]
  \centering
  \subfigure[]{\includegraphics[height=3cm]{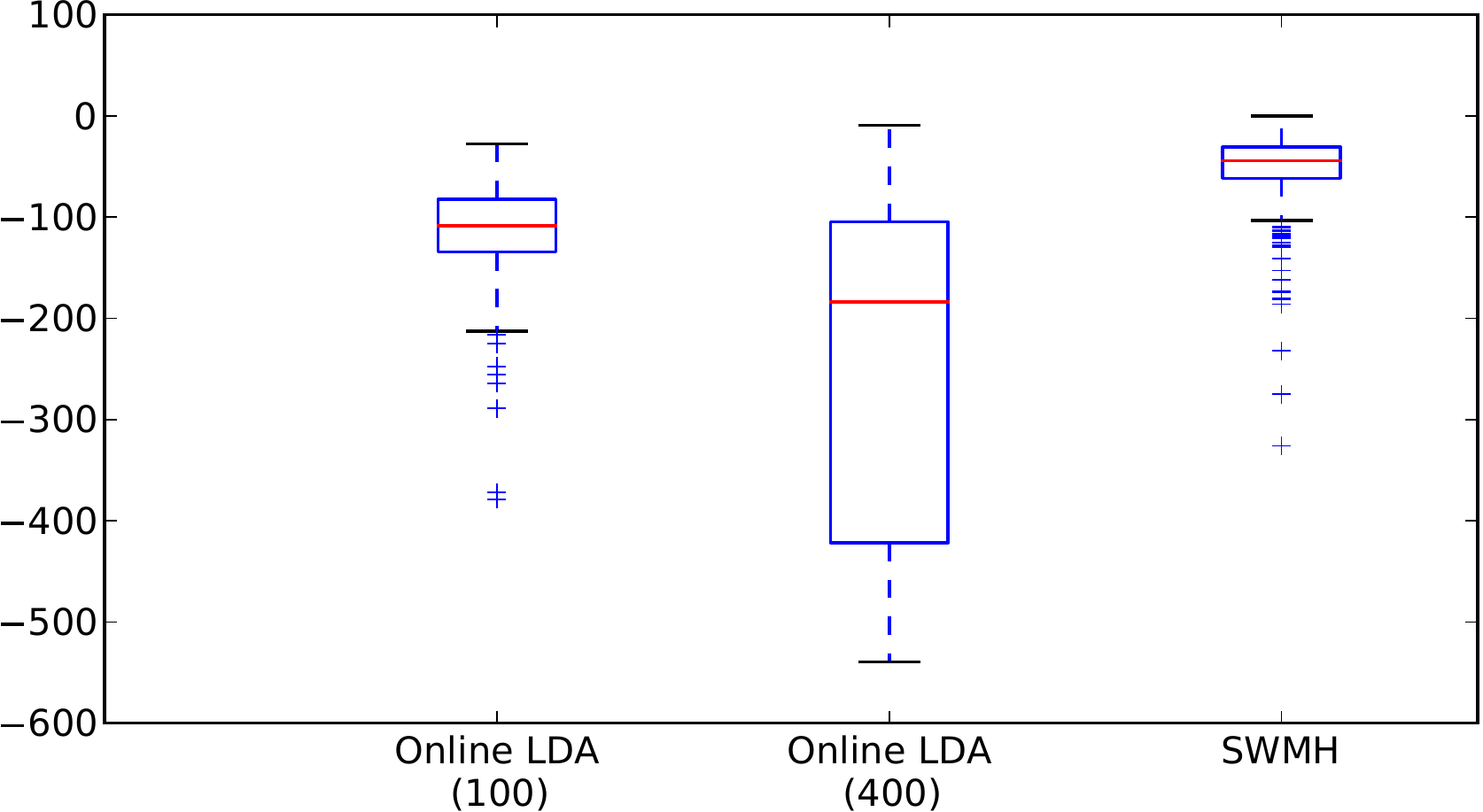}}\hspace{1cm}
  \subfigure[]{\includegraphics[height=3cm]{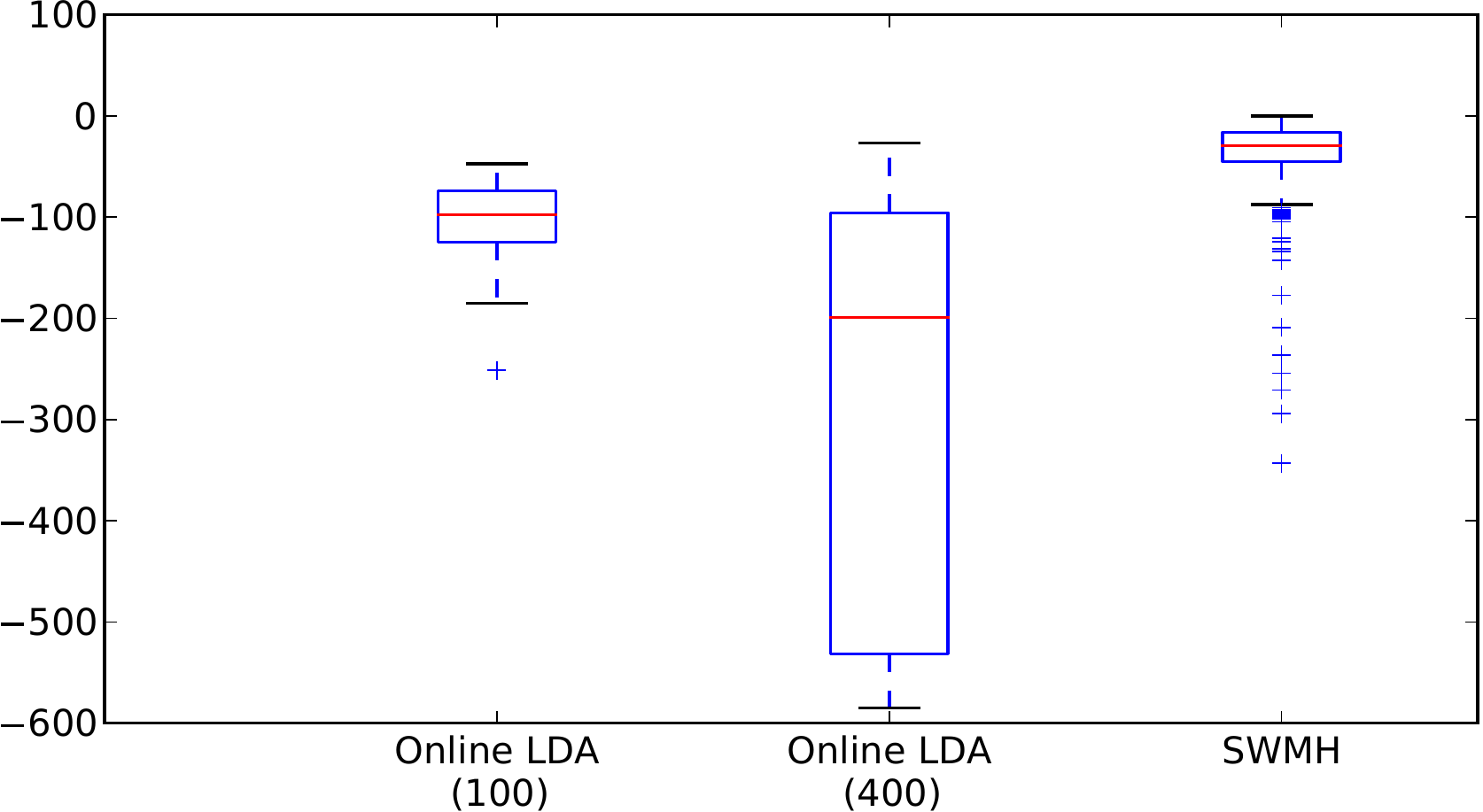}}
  \caption[]{Coherence of topics mined by SWMH vs Online LDA topics in the (a) 20 Newsgroups and (b) Reuters corpora.}
  \label{fig:com}
\end{figure}

In this evaluation we compare the quality of the topics mined by SWMH against Online LDA topics for the 20 Newsgroups and Reuters corpora. For this  
we measure \emph{topic coherence}, which is defined as
\[
C(t) = \sum\limits_{m=2}^{M} \sum\limits_{l=1}^{m-1} 
\log\frac{D(v_m,v_l)}{D(v_l)},
\]

\noindent where $D(v_l)$ is the document frequency of the term $v_l$, and 
$D(v_m,v_l)$ is the co-document frequency of the terms $v_m$ and 
$v_l$ \cite{coherence}.
This metric depends on the first $M$ elements of the topics. For our 
evaluations we fixed $M$ to $10$. However, we remark that the comparison is not direct since 
both the SWMH and Online LDA topics are different in nature: SWMH topics are subsets of the vocabulary 
with uniform distributions while Online LDA topics are distributions over the complete vocabulary. 
In addition, Online LDA generates a fixed number of topics which is in the hundreds while SWMH produces thousands of topics. 
For the comparison we chose the $n$-best mined topics by ranking them using an ad hoc metric involving the co-occurrence of the first element of the topic.
For the purpose of the evaluation we limited the SWMH to the $500$ best ranked topics.  
Figure~\ref{fig:com} shows the coherence for each corpus. In general, 
we can see a difference in the shape and quality of the coherence box plots. 
However, we notice that SWMH produces a considerable amount of outliers, which calls for further research in the ranking of the mined topics and their relation with the coherence. 

\section{Discussion and Future Work}
\label{sec:concl}
In this work we presented a large-scale approach to automatically mine topics in a given corpus based on Sampled Weighted Min-Hashing. The mined topics consist of subsets of highly correlated terms from the vocabulary. The proposed approach is able to mine topics in corpora which go from the thousands of documents ($1$ min approx.) to the millions of documents ($7$ hrs. approx.), including topics similar to the ones produced by Online LDA. We found that the mined topics can be used to represent a document for classification. 
We also showed that the complexity of the proposed approach grows linearly with the amount of documents.  
Interestingly, some of the topics mined by SWMH are related to the structure of the documents (e.g., in 
NIPS the words in the first topic correspond to parts of an article) and others 
to specific groups (e.g., team sports in 20 Newsgroups and Reuters, or the 
\emph{Transformers} universe in Wikipedia).  These examples suggest that SWMH 
is able to generate topics at different levels of granularity.

Further work has to be done to make sense of overly specific topics or to filter them out. In this direction, we found that weighting the terms has the effect of discarding several irrelevant topics and producing more compact ones. Another alternative, it is to restrict the vocabulary to the top most frequent terms as done by other approaches. Other interesting future work include exploring other weighting schemes, finding a better representation of documents from the mined topics and parallelizing SWMH.

\bibliographystyle{plainnat}
\bibliography{references}
\end{document}